\documentclass[letterpaper, 10 pt, conference]{ieeeconf}  

\IEEEoverridecommandlockouts                              

\usepackage{amsmath} 
\usepackage{amssymb}

\usepackage{ltl} 

\usepackage[vlined]{algorithm2e}

\usepackage{color}
\usepackage{subfig}
\usepackage{tikz}
\usepackage{hyperref}
\usetikzlibrary{arrows,fit,shapes,automata}
\usetikzlibrary{positioning,fit,calc,shapes}
\usetikzlibrary{decorations.fractals}
\usetikzlibrary{decorations.markings}

\newcommand{\comment}[1]{}

\newtheorem{example}{Example}
\newtheorem{theorem}{Theorem}
\newcommand*{\qed}{\hfill\ensuremath{\blacksquare}}

\newcommand{\init}{\mathsf{init}}
\newcommand{\belief}{\mathsf{belief}}
\newcommand{\abstr}{\mathsf{abstract}}
\newcommand{\vis}{\mathit{vis}}
\newcommand{\succs}{\mathit{succ}}
\newcommand{\beliefs}{\mathcal{P}(L_t)}

\newcommand{\states}{S}
\newcommand{\trans}{T}
\newcommand{\part}{\mathcal{Q}}

\newcommand{\post}{\mathit{succ}_t}

\newcommand{\outcome}{\mathit{outcome}}
\newcommand{\counterex}{\mathcal{C}}

\newcommand{\bools}{\mathbb{B}}
\newcommand{\true}{\mathit{true}}
\newcommand{\false}{\mathit{false}}
\newcommand{\nats}{\mathbb{N}}

\newcommand{\SP}{\mathcal{SP}}
\newcommand{\AP}{\mathcal{AP}}

\title{\LARGE \bf Synthesis of Surveillance Strategies via Belief Abstraction}

\author{Suda Bharadwaj$^{1}$ and Rayna Dimitrova$^{2}$ and Ufuk Topcu$^{1}$
\thanks{$^{1}$Suda Bharadwaj and Ufuk Topcu are with the University of Texas at Austin}%
\thanks{$^{2}$Rayna Dimitrova is with the University of Leicester, UK. Most of this work was done while Rayna Dimitrova was a postdoctoral researcher at UT Austin.}%
}

\begin{document}

\maketitle
\thispagestyle{empty}
\pagestyle{empty}


\begin{abstract}
We study the problem of synthesizing a controller for a robot with a \emph{surveillance objective}, that is, the robot is required to  maintain knowledge of the location of a moving, possibly adversarial target. We formulate this problem as a one-sided partial-information  game in which the winning condition for the agent is specified as a temporal logic formula. The specification formalizes the surveillance requirement given by the user, including additional non-surveillance tasks. In order to synthesize a surveillance strategy that meets the specification, we transform the partial-information game into a perfect-information one, using abstraction to mitigate the exponential blow-up typically incurred by such transformations. This enables the use of off-the-shelf tools for reactive synthesis. We use counterexample-guided refinement to automatically achieve abstraction precision that is sufficient to synthesize a surveillance strategy. We evaluate the proposed method on two case-studies, demonstrating its applicability to large state-spaces and diverse requirements.\looseness=-1

\end{abstract}


\section{INTRODUCTION}

Performing surveillance, that is, tracking the location of a target, has many applications. If the target is adversarial, these applications include patrolling and defense, especially in combination with other objectives, such as providing certain services or accomplishing a mission. Techniques for tracking non-adversarial but unpredictable targets have been proposed in settings like surgery to control cameras to keep a patient's organs under observation despite unpredictable motion of occluding obstacles \cite{Bandyopadhyay2011}. Mobile robots in airports have also been proposed to carry luggage for clients, requiring the robots to follow the human despite unpredictable motion and possibly sporadically losing sight of the target \cite{GonBanos02}. \looseness=-1

When dealing with a possibly adversarial target, a strategy for the surveying agent for achieving its objective can be seen as a strategy in a two-player game between the agent and the target. Since the agent may not always observe, or even know, the exact location of the target, surveillance is, by its very nature, a partial-information problem.
It is thus natural to reduce surveillance strategy synthesis to computing a winning strategy for the agent in a two-player partial-information game. Game-based models for related problems have been extensively studied in the literature. Notable examples include pursuit-evasion games~\cite{Chung2011}, patrolling games~\cite{Basilico12}, and graph-searching games~\cite{Kreutzer11}, where the problem is formulated as enforcing eventual detection, which is, in its essence a search problem -- once the target is detected, the game ends. For many applications, this formulation is too restrictive. Often, the goal is not to detect or capture the target, but to maintain certain level of information about its location over an unbounded (or infinite) time duration, or, alternatively, be able to obtain sufficiently precise information over and over again. In other cases, the agent has an additional objective, such as performing certain task, which might prevent him from capturing the target, but allow for satisfying a more relaxed surveillance objective.

In this paper, we study the problem of synthesizing strategies for enforcing \emph{temporal surveillance objectives}, such as the requirement to never let the agent's uncertainty about the target's location exceed a given threshold, or recapturing the target every time it escapes. To this end, we consider surveillance objectives specified in linear temporal logic (LTL), equipped with basic surveillance predicates. This formulation also allows for a seamless combination with other task specifications. Our computational model is that of a two-player game played on a finite graph, whose nodes represent the possible locations of the agent and the target, and whose edges model the possible (deterministic) moves between locations. The agent plays the game with partial information, as it can only observe the target when  it is in it's area of sight. The target, on the other hand, always has full information about the agent's location, even when the agent is not in sight. In that way, we consider a model with one-sided partial information, making the computed strategy for the agent robust against a potentially more powerful adversary. \looseness=-1

We formulate surveillance strategy synthesis as the problem of computing a winning strategy for the agent in a partial-information game with a surveillance objective. There is a rich theory on partial-information games with LTL objectives~\cite{DoyenR11,Chatterjee2013}, and it is well known that even for very simple objectives the synthesis problem is EXPTIME-hard~\cite{Reif84,BerwangerD08}. Moreover, all the standard algorithmic solutions to the problem are based on some form of \emph{belief set construction}, which transforms the imperfect-information game into a perfect-information game and this may be exponentially larger, since the new set of states is the powerset of the original one. Thus, such approaches scale poorly in general, and are not applicable in most practical situations.

We address this problem by using \emph{abstraction}. We introduce an \emph{abstract belief set construction}, which underapproximates the information-tracking abilities of the agent (or, alternatively, overapproximates its belief, i.e., the set of positions it knows the target could be in). Using this construction we reduce surveillance synthesis to a two-player perfect-information game with LTL objective, which we then solve using off-the shelf reactive synthesis tools~\cite{EhlersR16}. Our construction guarantees that the abstraction is sound, that is, if a surveillance strategy is found in the abstract game, it corresponds to a surveillance strategy for the original game. If, on the other hand, such a strategy is not found, then the method automatically checks if this is due to the coarseness of the abstraction, in which case the abstract belief space is automatically refined. Thus, our method follows the general counterexample guided abstraction refinement (CEGAR)~\cite{ClarkeGJLV00} scheme, which has successfully demonstrated its potential in formal verification and reactive synthesis.

{\bf Contributions.} We make the following contributions:\\
(1) We propose a \emph{formalization of surveillance objectives} as temporal logic specifications, and frame surveillance strategy synthesis  as a partial-information reactive synthesis problem.\\
(2) We develop an \emph{abstraction method that soundly approximates} surveillance strategy synthesis, thus enabling the application of efficient techniques for reactive synthesis.\\
(3) We design procedures that \emph{automatically refine a given abstraction} in order to improve its precision when no surveillance strategy exists due to coarseness of the approximation.\\
(4) We evaluate our approach on different surveillance objectives (e.g, safety, and liveness) combined with task specifications, and discuss the qualitatively different behaviour of the synthesized strategies for the different kinds of specifications.

{\bf Related work.}
While closely related to the surveillance problem we consider, pursuit-evasion games with partial information~\cite{Chung2011, Chin2010, Antoniades2003} formulate the problem as eventual detection, and do not consider combinations with other mission specifications. Other work, such as \cite{Vidal2002} and \cite{Kim2001}, additionally incorporates map building during pursuit in an unknown environment, but again solely for target detection.

Synthesis from LTL specifications~\cite{Pnueli1989}, especially from formulae in the efficient GR(1) fragment~\cite{Piterman2006}, has been extensively used in robotic planning (e.g.~\cite{wong2012,Kress2007}), but surveillance-type objectives, such as the ones we study here, have not been considered so far. Epistemic logic specifications~\cite{MeydenV98} can refer to the knowledge of the agent about the truth-value of logical formulas, but, contrary to our surveillance specifications, are not capable of expressing requirements on the size of the agent's uncertainty.

CEGAR has been developed for verification~\cite{ClarkeGJLV00}, and later for control~\cite{HenzingerJM03}, of LTL specifications. 
It has also been extended to infinite-state partial-information games~\cite{DimitrovaF08}, and used for sensor design~\cite{FuDT14}, both in the context of safety specifications. In addition to being focused on safety objectives, the refinement method in~\cite{DimitrovaF08} is designed to provide the agent with just enough information to achieve safety, and is thus not applicable to surveillance properties whose satisfaction depends on the size of the belief sets.

\section{GAMES WITH SURVEILLANCE OBJECTIVES}
We begin by defining a formal model for describing surveillance strategy synthesis problems, in the form of a two-player game between an agent and a target, in which the agent has only partial information about the target's location.

\subsection{Surveillance Game Structures}\label{sec:surveillance-games}
We define a \emph{surveillance game structure} to be  a tuple $G  = (\states,s^\init,\trans,\vis)$, with the following components:
\begin{itemize}
\item $\states = L_a \times L_t$ is the set of states, with $L_a$ the set of locations of the agent, and $L_t$ the locations of the target;
\item $s^\init = (l_a^\init,l_t^\init)$ is the initial state;
\item $\trans \subseteq \states \times \states$ is the transition relation describing the possible moves of the agent and the target; and
\item $\vis : \states \to \bools$ is a function that maps a state $(l_a,l_t)$ to $\true$ iff \emph{ position $l_t$ is in the area of sight of $l_a$}.
\end{itemize}

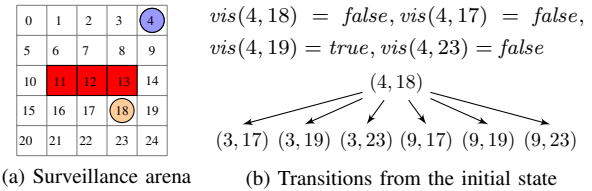
\begin{figure}
\subfloat[Surveillance arena \label{simple-grid}]{
\begin{tikzpicture}[scale=0.8]
\draw[step=0.5cm,color=gray] (-1.5,-1.5) grid (1,1);
\filldraw[fill=blue,draw=black] (+0.75,+0.75) circle (0.2cm);
\filldraw[fill=red,draw=black] (0,0) rectangle (-0.5,-0.5);
\filldraw[fill=red,draw=black] (-0.5,0) rectangle (-1,-0.5);
\filldraw[fill=red,draw=black] (0,0) rectangle (0.5,-0.5);
\filldraw[fill=blue!40!white,draw=black] (+0.75,+0.75) circle (0.2cm);
\filldraw[fill=orange!40!white,draw=black] (0.25,-0.75) circle (0.2cm);
\node at (-1.30,+0.75) {\tiny{0}};
\node at (-0.80,+0.75) {\tiny{1}};
\node at (-0.30,+0.75) {\tiny{2}};
\node at (0.20,+0.75) {\tiny{3}};
\node at (0.73,+0.75) {\tiny{4}};
\node at (-1.33,+0.25) {\tiny{5}};
\node at (-0.85,+0.25) {\tiny{6}};
\node at (-0.35,+0.25) {\tiny{7}};
\node at (0.25,+0.25) {\tiny{8}};
\node at (0.75,+0.25) {\tiny{9}};
\node at (-1.28,-0.27) {\tiny{10}};
\node at (-0.78,-0.27) {\tiny{11}};
\node at (-0.28,-0.27) {\tiny{12}};
\node at (0.28,-0.27) {\tiny{13}};
\node at (0.75,-0.25) {\tiny{14}};
\node at (-1.3,-0.75) {\tiny{15}};
\node at (-0.8,-0.75) {\tiny{16}};
\node at (-0.3,-0.75) {\tiny{17}};
\node at (0.25,-0.75) {\tiny{18}};
\node at (0.75,-0.75) {\tiny{19}};
\node at (-1.35,-1.25) {\tiny{20}};
\node at (-0.85,-1.25) {\tiny{21}};
\node at (-0.35,-1.25) {\tiny{22}};
\node at (0.25,-1.25) {\tiny{23}};
\node at (0.75,-1.25) {\tiny{24}};
\end{tikzpicture}
\hspace{.3cm}}
\subfloat[Transitions from the initial state\label{fig:simple-transitions}]{
\begin{minipage}{5.0cm}
\vspace{-1.8cm}
{\fontsize{8}{10}\selectfont $\vis(4,18) = \false,\vis(4,17) = \false,$ $\vis(4,19) = \true, \vis(4,23) = \false$}

\smallskip

\begin{tikzpicture}[node distance=.75 cm,auto,>=latex',line join=bevel,transform shape,scale=0.75]
\node at (0,0) (s0) {$(4,18)$};
\node  [below left of=s0,yshift=-.5cm] (s3) {$(3,23)$};
\node  [below right of=s0,yshift=-.5cm] (s4) {$(9,17)$};
\node  [left of=s3,xshift=-.35cm] (s2) {$(3,19)$};
\node  [left of=s2,xshift=-.35cm] (s1) {$(3,17)$};
\node  [right of=s4,xshift=.35cm] (s5) {$(9,19)$};
\node  [right of=s5,xshift=.35cm] (s6) {$(9,23)$};

\draw [->] (s0) edge (s1.north);
\draw [->] (s0) edge (s2.north);
\draw [->] (s0) edge (s3.north);
\draw [->] (s0) edge (s4.north);
\draw [->] (s0) edge (s5.north);
\draw [->] (s0) edge (s6.north);
\end{tikzpicture}
\end{minipage}
}
\vspace{-0.1cm}
\caption{A simple surveillance game on a grid arena. Obstacles are shown in red, the agent (at location 4) and the target (at location 18) are coloured in blue and orange respectively.}
\label{fig:simple-surveillance-game}
\vspace{-.7cm}
\end{figure}

The transition relation $T$ encodes the one-step move of both the target and the agent, where the target moves first and the agent moves second. For a state $(l_a,l_t)$ we denote $\succs_t(l_a,l_t)$ as the set of successor locations of the target:

$\succs_t(l_a,l_t) = \{l_t' \in L_t \mid \exists l_a'.\ ((l_a,l_t),(l_a',l_t')) \in T\}$.

We extend $\succs_t$ to sets of locations of the target by stipulating that the set $\post(l_a,L)$ consists of all possible successor locations of the target for states in $\{l_a\} \times L$. Formally, let $\post(l_a, L) = \bigcup_{l_t \in L}\succs_t(l_a,l_t)$.

For a state $(l_a,l_t)$ and a successor location of the target $l_t'$, we denote with $\succs_a(l_a,l_t,l_t')$ the set of successor locations of the agent, given that the target moves to $l_t'$: 

$\succs_a(l_a,l_t,l_t') = \{l_a' \in L_a \mid  ((l_a,l_t),(l_a',l_t')) \in T\}$.

We assume that, for every $s \in \states$, there exists $s' \in \states$ such that $(s,s') \in T$, that is, from every state there is at least one move possible (this might be staying in the same state). We also assume that when the target moves to an invisible location, its position does not influence the possible one-step moves of the agent. Formally, we require that if $\vis(l_a,l_t''') = \vis(l_a,l_t'''')=\false$, then $\succs_a(l_a,l_t',l_t''') = \succs_a(l_a,l_t'',l_t'''')$ for all $l_t',l_t'',l_t''',l_t'''' \in L_t$. This assumption is natural in the setting when the agent can move in one step only to locations that are in its sight.

\begin{example}\label{ex:simple-surveillance-game}
Figure~\ref{fig:simple-surveillance-game} shows an example of a surveillance game on a grid.  The sets of possible locations $L_a$ and $L_t$ for the agent and the target consist of the squares of the  grid. The transition relation $T$ encodes the possible one-step moves of both the agent and the target on the grid, and incorporates all desired constraints. For example, moving to an occupied location, or an obstacle, is not allowed. Figure~\ref{fig:simple-transitions} shows the possible transitions from the initial state $(4,18)$.

The function $\vis$ encodes straight-line visibility: a location $l_t$ is visible from a location $l_a$ if there is no obstacle on the straight line between them. Initially the target is not in the area of sight of the agent, but the agent knows the initial position of the target. However, once the target moves to one of the locations reachable in one step, in this case, locations $\{17,19,23\}$, this might no longer be the case. More precisely, if the target moves to location $19$, then the agent observes its location, but if it moves to one of the others, then the agent no longer knows its exact location. \qed
\end{example}

\subsection{Belief-Set Game Structures}

In surveillance strategy synthesis we need to state properties of, and reason about, the information which the agent has, i.e. its \emph{belief} about the location of the target. To this end, we can employ a powerset construction which is commonly used to transform a partial-information game into a perfect-information one, by explicitly tracking the knowledge one player has as a set of possible states of the other player.

Given a set $B$, we denote with $\mathcal{P}(B) = \{B' \mid B'\subseteq B\}$ the powerset (set of all subsets) of $B$.

For a surveillance game structure $G  = (\states,s^\init,\trans,\vis)$ we define the corresponding \emph{belief-set game structure} $G_\belief  = (\states_\belief,s^\init_\belief,\trans_\belief)$ with the following components:
\begin{itemize}
\item $\states_\belief = L_a \times \beliefs$ is the set of states, with $L_a$ the set of locations of the agent, and $\beliefs$ the set of \emph{belief sets} describing information about the location of the target;
\item $s^\init_\belief = (l_a^\init,\{l_t^\init\})$ is the initial state;
\item $\trans_\belief \subseteq \states_\belief \times \states_\belief$ is the transition relation where $((l_a, B_t),(l_a', B_t')) \in \trans_\belief$ iff $l_a' \in  \succs_a(l_a,l_t,l_t')$ for some $l_t \in B_t$ and $l_t' \in B_t'$ and one of these holds:
\begin{itemize}
\item[(1)] $B_t' = \{l_t'\}$, $l_t' \in \post(l_a,B_t)$, $\vis(l_a,l_t') = \true$;
\item[(2)] $B_t' = \{l_t' \in \post(l_a,B_t)  \mid  \vis(l_a,l_t') = \false \}$.
\end{itemize}
\end{itemize}
Condition (1) captures the successor locations of the target that can be observed from the agent's current position $l_a$. Condition (2) corresponds to the belief set consisting of \emph{all possible successor locations of the target not visible from $l_a$}. 

\begin{figure}
\begin{center}
\begin{tikzpicture}[node distance=.9 cm,auto,>=latex',line join=bevel,transform shape,scale=.75]
\node at (0,0) (s0) {$(4,\{18\})$};
\node  [below left of=s0,yshift=-.5cm,xshift=-.35cm] (s2) {$(3,\{17,23\})$};
\node  [below right of=s0,yshift=-.5cm,xshift=.35cm] (s3) {$(9,\{19\})$};
\node  [left of=s2,xshift=-1cm] (s1) {$(3,\{19\})$};
\node  [right of=s3,xshift=1cm] (s4) {$(9,\{17,23\})$};

\draw [->] (s0) edge (s1.north);
\draw [->] (s0) edge (s2.north);
\draw [->] (s0) edge (s3.north);
\draw [->] (s0) edge (s4.north);
\end{tikzpicture}
\end{center}

\vspace{-.3cm}
\caption{Transitions from the initial state in the belief-set game from Example~\ref{ex:simple-belief-game} where $\vis(4,17) = \vis(4,23) = \false$.}
\label{fig:simple-belief-game}
\vspace{-.5cm}
\end{figure}
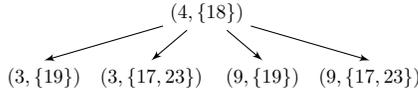

\begin{example}\label{ex:simple-belief-game}
Consider the surveillance game structure from Example~\ref{ex:simple-surveillance-game}. The initial belief set is $\{18\}$, consisting of the target's initial position. After the first move of the target, there are two possible belief sets: the set $\{19\}$ resulting from the move to a location in the area of sight of the agent, and $\{17,23\}$ consisting of the two invisible locations reachable in one step from location $18$.
Figure~\ref{fig:simple-belief-game} shows the successor states of the initial state $(4,\{18\})$ in $G_\belief$. \qed
\end{example}

Based on  $T_\belief$, we can define the functions $\succs_t : \states_\belief \to \mathcal{P}(\beliefs)$ and  $\succs_a : \states_\belief \times \beliefs \to \mathcal{P}(L_a)$ similarly to the corresponding functions defined for $G$. 

A \emph{run} in $G_\belief$ is an infinite sequence $s_0,s_1,\ldots$ of states in $\states_\belief$, where $s_0 = s_\belief^\init$,  $(s_i,s_{i+1}) \in T_\belief$ for all $i$. 

A \emph{strategy for the target in $G_\belief$} is a function $f_t: \states_\belief^+ \to \beliefs$ such that $f_t(\pi\cdot s) = B_t$ implies $B_t \in \succs_t(s)$ for every $\pi \in \states_\belief^*$ and $s \in \states_\belief$. That is, a strategy for the target suggests a move resulting in some belief set reachable from some location in the current belief.

A \emph{strategy for the agent in $G_\belief$} is a function $f_a : S^+ \times \beliefs \to S$ such that $f_a(\pi\cdot s,B_t) = (l_a',B_t')$ implies $B_t' = B_t$ and $l_a' \in \succs_a(s,B_t)$ for every $\pi \in \states_\belief^*$, $s \in \states_\belief$ and $B_t \in \beliefs$. Intuitively, a strategy for the agent suggests a move based on the observed history of the play and the current belief about the target's position.

The outcome of given strategies $f_a$ and $f_t$ for the agent and the target in $G_\belief$, denoted $\outcome(G_\belief,f_a,f_t)$, is a run $s_0,s_1,\ldots$ of $G_\belief$ such that for every $i \geq 0$, we have $s_{i+1} = f_a(s_0,\ldots,s_i,B_t^i)$, where $B_t^i = f_t(s_0,\ldots,s_i)$.

\subsection{Temporal Surveillance Objectives}
Since the states of a belief-set game structure track the information that the agent has, we can state and interpret surveillance objectives over its runs. We now formally define the surveillance properties in which we are interested. 

We consider a set of \emph{surveillance predicates} $\SP = \{p_k \mid k \in \nats_{>0}\}$, where for $k \in \nats_{>0}$ we say that a state $(l_a,B_t)$ in the belief game structure satisfies $p_k$ (denoted $(l_a,B_t) \models p_k$) iff 
$|\{l_t \in B_t \mid \vis(l_a,l_t)  = \false \}| \leq k$. Intuitively, $p_k$ is satisfied by the states in the belief game structure where the size of the belief set does not exceed the threshold $k \in \nats_{>0}$.

We study surveillance objectives expressed by formulas of linear temporal logic (LTL) over surveillance predicates.
 The LTL surveillance formulas  are generated by the grammar\\
$\varphi := p \mid \true \mid \false \mid \varphi \wedge \varphi \mid \varphi \vee \varphi \mid \LTLnext  \varphi  \mid \varphi \LTLuntil \varphi \mid \varphi \LTLrelease \varphi,$\\
where $p \in \SP$ is a surveillance predicate, $\LTLnext$ is the \emph{next} operator, $\LTLuntil$ is the \emph{until} operator, and $\LTLrelease$ is the \emph{release} operator. We also define the derived operators 
\emph{finally}: $\LTLfinally \varphi = \true \LTLuntil \varphi$ and 
\emph{globally}: $\LTLglobally \varphi = \false \LTLrelease \varphi$.

LTL formulas are interpreted over (infinite) runs. If a run $\rho$ satisfies an LTL formula $\varphi$, we write $\rho \models \varphi$. The formal definition of LTL semantics can be found in~\cite{BaierKatoen08}. Here we informally explain the meaning of the formulas we use.

Of special interest will be surveillance formulas of the form $\LTLglobally p_k$, termed \emph{safety surveillance objective}, and $\LTLglobally\LTLfinally p_k$, called \emph{liveness surveillance objective}.
Intuitively, the safety surveillance formula $\LTLglobally p_k$ is satisfied if at each point in time the size of the belief set does not exceed $k$. The liveness surveillance objective $\LTLglobally\LTLfinally p_k$, on the other hand, requires that infinitely often this size is below or equal to $k$.

\begin{example}
We can specify that the agent is required to always know with certainty the location of the target as
$\LTLglobally p_1$.
A more relaxed requirement is that the agent's uncertainty never grows above $5$ locations, and it infinitely often reduces this uncertainty to at most $2$ locations: $\LTLglobally p_5 \wedge \LTLglobally\LTLfinally p_2$.
\qed
\end{example}

\subsection{Incorporating Task Specifications}
We can integrate LTL objectives not related to surveillance, i.e., \emph{task specifications}, by considering, in addition to $\SP$, a set $\AP$ of atomic predicates interpreted over states of $G$. In order to define the semantics of $p \in \AP$ over states of $G_\belief$, we restrict ourselves to predicates observable by the agent. 
Formally, we require that for $p \in \AP$, and states $(l_a,l_t')$ and $(l_a,l_t'')$ with $\vis(l_a,l_t')=\vis(l_a,l_t'')=\false$ it holds that $(l_a,l_t') \models p$ iff $(l_a,l_t'') \models p$. One class of such predicates are those that depend only on the agent's position.

\begin{example}
Suppose that $\mathit{at\_goal}$ is a predicate true exactly when the agent is at some designated goal location. We can then state that the agent visits the goal infinitely often while always maintaining belief uncertainty of at most $10$ locations using the LTL formula $\LTLglobally\LTLfinally \mathit{at\_goal} \wedge \LTLglobally p_{10}$.
\qed
\end{example}

\subsection{Surveillance Synthesis Problem}
A \emph{surveillance game} is a pair $(G,\varphi)$, where $G$ is a surveillance game structure and $\varphi$ is a surveillance objective. A \emph{winning strategy for the agent for $(G,\varphi)$} is a strategy $f_a$ for the agent in the corresponding belief-set game structure $G_\belief$ such that for every strategy $f_t$ for the target in $G_\belief$ it holds that $\outcome(G_\belief,f_a,f_t) \models \varphi$. Analogously, a \emph{winning strategy for the target for $(G,\varphi)$} is a strategy $f_t$ such that, for every strategy $f_a$ for the agent in $G_\belief$, it holds that $\outcome(G_\belief,f_a,f_t) \not\models \varphi$.

{\bf Surveillance synthesis problem:} Given a surveillance game $(G,\varphi)$, compute a winning strategy for the agent for $(G,\varphi)$, or determine that such a strategy does not exist.

It is well-known that two-player perfect-information games with LTL objectives over finite-state game structures are determined, that is exactly one of the players has a winning strategy. This means that the agent does not have a winning strategy for a given surveillance game, if and only if the target has a winning strategy for this game. We refer to winning strategies of the target as \emph{counterexamples}.


\section{BELIEF SET ABSTRACTION}
We used the belief-set game structure in order to state the surveillance objective of the agent. While in principle it is possible to solve the surveillance strategy synthesis problem on this game, this is in most cases computationally infeasible, since the size of this game is exponential in the size of the original game. To circumvent this construction when possible, we propose an abstraction-based method, that given a surveillance game structure and a partition of the set of the target's locations, yields an approximation that is sound with respect to surveillance objectives for the agent.

An \emph{abstraction partition} is a family $\part = \{Q_i\}_{i=1}^n$ of subsets of $L_t$, $Q_i \subseteq L_t$ such that the following hold:
\begin{itemize}
\item $\bigcup_{i=1}^n Q_i = L_t$ and $Q_i \cap Q_j = \emptyset$ for all $i \neq j$;
\item For each $p \in \AP$, $Q \in \part$ and $l_a \in L_a$, it holds that $(l_a,l_t') \models p$ iff $(l_a,l_t'') \models p$ for every $l_t',l_t'' \in Q$.
\end{itemize}
Intuitively, these conditions mean that $\mathcal Q$ partitions the set of locations of the target, and the concrete locations in each of the sets in $\part$ agree on the value of the  propositions in $\AP$.

If $\part' =  \{Q_i'\}_{i=1}^m$ is a family of subsets of $L_t$ such that $\bigcup_{i=1}^m Q_i' = L_t$ and for each $Q_i' \in \part'$ there exists $Q_j \in \part$ such that $Q_i' \subseteq Q_j$, then $\part'$ is also an abstraction partition, and we say that $\part'$ \emph{refines} $\part$, denoted $\part' \preceq \part$.

For $\part = \{Q_i\}_{i=1}^n$,  we define a function $\alpha_\part : L_t \to \part$ by $\alpha(l_t) = Q$ for the unique $Q \in \part$ with $l_t \in Q$. We denote also with $\alpha_{\part} : \mathcal{P}(L_t) \to \mathcal{P}(\part)$ the \emph{abstraction function} defined by $\alpha_{\part}(L) = \{\alpha_\part(l) \mid l \in L\}$.
We define a \emph{concretization function} $\gamma :  \mathcal{P}(\part) \cup L_t \to \mathcal{P}(L_t)$ such that 
$\gamma(A) = \{l_t\}$ if $A = l_t \in L_t$, and  $\gamma(A) = \bigcup_{Q \in A} Q$ if $A \in \mathcal{P}(\part)$.

Given a surveillance game structure $G  = (\states,s^\init,\trans,\vis)$ and an abstraction partition $\part = \{Q_i\}_{i=1}^n$ of the set $L_t$, we define the \emph{abstraction of $G$ w.r.t.\ $\part$} to be the game structure 
$G_\abstr  = \alpha_{\part}(G)= (\states_\abstr,s^\init_\abstr,\trans_\abstr)$, where

\begin{itemize}
\item $\states_\abstr = (L_a \times \mathcal P(\part)) \cup (L_a \times L_t)$  is the set of \emph{abstract states}, consisting of states approximating the belief sets in the game structure $G_\belief$, as well as the states $\states$;
\item $s^\init_\abstr = (l_a^\init,l_t^\init)$ is the \emph{initial abstract state};
\item $\trans_\abstr \subseteq \states_\abstr \times \states_\abstr$ is the transition relation such that $((l_a, A_t),(l_a', A_t')) \in \trans_\abstr$ if and only if one of the following two conditions is satisfied:
\begin{itemize}
\item[(1)] $A_t' = l_t'$, $l_t' \in \post(\gamma(A_t))$ and $\vis(l_a,l_t') = \true$, and
$l_a' \in \succs_a(l_a,l_t,l_t')$ for some $l_t \in \gamma(A_t)$.
\item[(2)] $A_t' = \alpha_{\part}(\{l_t' \in \post(\gamma(A_t))  |  \vis(l_a,l_t') = \false\})$, and
$l_a' \in \succs_a(l_a,l_t,l_t')$ for some $l_t \in \gamma(A_t)$ and some
$l_t' \in \post(\gamma(A_t))$ with $\vis(l_a,l_t') = \false$.


\end{itemize}
\end{itemize}

As for the belief-set game structure, the first condition captures the successor locations of the target, which can be observed from the agent's current location $l_a$. Condition (2) corresponds to the \emph{abstract belief set} whose concretization  consists of all possible successors of all positions in $\gamma(A_t)$, which are  not visible from $l_a$. Since the belief abstraction overapproximates the agent's belief, that is, $\gamma(\alpha_{\part}(B)) \supseteq B$, the next-state abstract belief $\gamma(A_t')$ may include positions in $L_t$ that are not successors of positions in $\gamma(A_t)$.

%
%
%

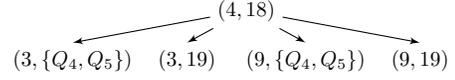
\begin{figure}
\begin{center}
\begin{tikzpicture}[node distance=.9 cm,auto,>=latex',line join=bevel,transform shape,scale=.8]
\node at (0,0) (s0) {$(4,18)$};
\node  [below left of=s0,yshift=-.2cm,xshift=-.35cm] (s2) {$(3,19)$};
\node  [below right of=s0,yshift=-.2cm,xshift=.35cm] (s3) {$(9,\{Q_4,Q_5\})$};
\node  [left of=s2,xshift=-1cm] (s1) {$(3,\{Q_4,Q_5\})$};
\node  [right of=s3,xshift=1cm] (s4) {$(9,19)$};

\draw [->] (s0) edge (s1.north);
\draw [->] (s0) edge (s2.north);
\draw [->] (s0) edge (s3.north);
\draw [->] (s0) edge (s4.north);
\end{tikzpicture}
\end{center}
\vspace{-.2cm}
\caption{Transitions from the initial state in the abstract game from Example~\ref{ex:simple-abstr-game} where $\alpha_\part(17) = Q_4$ and $\alpha_\part(23) = Q_5$.}
\label{fig:simple-abstr-game}
\vspace{-.7cm}
\end{figure}

\begin{example}\label{ex:simple-abstr-game}
Consider again the surveillance game from Example~\ref{ex:simple-surveillance-game}, and the abstraction partition $\part = \{Q_1,\ldots,Q_5\}$, where the set $Q_i$ corresponds to the $i$-th row of the grid. For location $17$ of the target we have $\alpha_\part(17) = Q_4$, and for  $23$ we have $\alpha_\part(23) = Q_5$. Thus, the belief set $B = \{17,23\}$ is covered by the abstract belief set $\alpha_Q(B) = \{Q_4,Q_5\}$. Figure~\ref{fig:simple-abstr-game} shows the successors of the initial state $(4,18)$ of the abstract belief-set game structure. The concretization of the abstract belief set $\{Q_4,Q_5\}$ is the set $\{15,16,17,18,19,20,21,22,23,24\}$ of target locations.\qed
\end{example}

An abstract state $(l_a,A_t)$ \emph{satisfies a surveillance predicate $p_k$}, denoted $(l_a,A_t) \models p_k$, iff 
$|\{l_t \in \gamma(A_t) \mid \vis(l_a,l_t)  = \false \}| \leq k$. Simply, the number of states in the concretized belief set has to be less than or equal to $k$. Similarly, for a predicate $p \in \AP$, we define $(l_a,A_t) \models p$ iff for every $l_t \in \gamma(A_t)$ it holds that $(l_a,l_t) \models p$. With these definitions, we can interpret surveillance objectives over runs of $G_\abstr$.

Strategies (and wining strategies) for the agent and the target in an abstract belief-set game $(\alpha_\part(G),\varphi)$ are defined analogously to strategies (and winning strategies) in $G_\belief$.

In the construction of the abstract  game structure, we overapproximate the belief-set of the agent at each step. Since we consider surveillance predicates that impose upper bounds on the size of the belief, such an abstraction  gives more power to the target (and, dually less power to the agent).  This construction guarantees that the abstraction is \emph{sound}, meaning that an abstract strategy for the agent that achieves a surveillance objective corresponds to a winning strategy in the concrete game. This is stated in the following theorem.

\begin{theorem}
Let $G$ be a surveillance game structure, $\part = \{Q_i\}_{i=1}^n$ be an abstraction partition, and $G_\abstr = \alpha_\part(G)$. For every surveillance objective $\varphi$, if there exists a wining strategy for the agent in the abstract belief-set game $(\alpha_\part(G),\varphi)$, then there exists a winning strategy for the agent in the concrete surveillance game $(G,\varphi)$.
\end{theorem}


\section{BELIEF REFINEMENT FOR SAFETY}
\subsection{Counterexample Tree}
A winning strategy for the target in a game with safety surveillance objective can be represented as a tree. 
An \emph{abstract counterexample tree} $\counterex_\abstr$ for $(G_\abstr,\LTLglobally p_k)$ is a finite tree,  whose nodes are labelled with states in $\states_\abstr$ such that the following conditions are satisfied:
\begin{itemize}
\item The root node is labelled with the initial state $s_\abstr^\init$.
\item A node is labelled with an abstract state  which violates $p_k$ (that is, $s_\abstr$ where $s_\abstr \not\models p_k$) iff it is a leaf.
\item The tree branches according to all possible transition choices of the agent. Formally, if an internal node $v$ is labelled with $(l_a,A_t)$, then there is unique $A_t'$  such that: (1) $((l_a,A_t),(l_a',A_t')) \in \trans_\abstr$ for some $l_a' \in L_a$, and (2) for every $l_a' \in L_a$ such that $((l_a,A_t),(l_a',A_t')) \in \trans_\abstr$, there is a child $v'$ of $v$ labelled with $(l_a',A_t')$.
\end{itemize}

A \emph{concrete counterexample tree} $\counterex_\belief$ for $(G_\belief,\LTLglobally p_k)$ is a finite tree defined analogously to an abstract counterexample tree with nodes labelled with states in $\states_\belief$.

Due to the overapproximation of the belief sets, not every counterexample in the abstract game corresponds to a winning strategy for the target in the original game.

An abstract counterexample $\counterex_\abstr$ in $(G_\abstr,\LTLglobally p_k)$ is \emph{concretizable} if there exists a concrete counterexample 
tree $\counterex_\belief$ in $(G_\belief,\LTLglobally p_k)$, that differs from $\counterex_\abstr$ only in the node labels, and each node labelled with $(l_a,A_t)$ in $\counterex_\abstr$ has label $(l_a, B_t)$ in $\counterex_\belief$ for which $B_t \subseteq \gamma(A_t)$.

\begin{figure}
\subfloat[Abstract counterexample tree\label{fig:simple-safety-counterex-abstr}]{
\begin{tikzpicture}[node distance=.7 cm,auto,>=latex',line join=bevel,transform shape,scale=.75]
\node at (0,0) (s0) {$(4,18)$};
\node  [below left of=s0,yshift=-.5cm,xshift=-.5cm] (s1) {$(3,\{Q_4,Q_5\})$};
\node  [below right of=s0,yshift=-.5cm,xshift=.5cm] (s2) {$(9,\{Q_4,Q_5\})$};

\draw [->] (s0) edge (s1.north);
\draw [->] (s0) edge (s2.north);
\end{tikzpicture}\hspace{.5cm}
}
\hfill
\subfloat[Concrete counterexample tree\label{fig:simple-safety-counterex-concr}]{
\begin{tikzpicture}[node distance=.7 cm,auto,>=latex',line join=bevel,transform shape,scale=.75]
\node at (0,0) (s0) {$(4,18)$};
\node  [below left of=s0,yshift=-.5cm,xshift=-.5cm] (s1) {$(3,\{17,23\})$};
\node  [below right of=s0,yshift=-.5cm,xshift=.5cm] (s2) {$(9,\{17,23\})$};

\draw [->] (s0) edge (s1.north);
\draw [->] (s0) edge (s2.north);
\end{tikzpicture}\hspace{.5cm}
}
\caption{Abstract and corresponding concrete counterexample trees for the surveillance game in Example~\ref{ex:simple-safety-counterex}.}
\label{fig:simple-safety-counterex}
\vspace{-.5cm}
\end{figure}
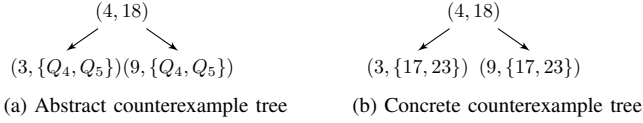

\begin{example}\label{ex:simple-safety-counterex}
Figure~\ref{fig:simple-safety-counterex-abstr} shows an abstract counterexample tree $\counterex_\abstr$ for the game $(\alpha_\part(G),\LTLglobally p_1)$, where $G$ is the surveillance game structure from Example~\ref{ex:simple-surveillance-game} and $\part$ is the abstraction partition from Example~\ref{ex:simple-abstr-game}. The counterexample corresponds to the choice of the target to move to one of the locations $17$ or $23$, which, for every possible move of the agent, results in an abstract state with abstract belief $B = \{Q_4,Q_5\}$ violating $p_1$.
A concrete counterexample tree $\counterex_\belief$ concretizing $\counterex_\abstr$ is shown in Figure~\ref{fig:simple-safety-counterex-concr}.
\qed
\end{example}

\subsection{Counterexample-Guided Refinement}
We now describe a procedure that determines whether an abstract counterexample for a safety surveillance objective is concretizable. This procedure essentially constructs the precise belief sets corresponding to the abstract moves of the target that constitute the abstract counterexample.

\subsubsection{Forward belief-set propagation}
Given an abstract counterexample tree $\counterex_\abstr$, we annotate its nodes with states in $\states_\belief$ in a top-down manner as follows. 
The root node is labelled with $s_\belief^\init$. 
If $v$ is a node annotated with the belief set $(l_a,B_t) \in \states_\belief$, and  $v'$ is a child of $v$ in $\counterex_\abstr$ labelled with an abstract state $(l_a',A_t')$, then we annotate $v'$ with the belief set $(l_a',B_t')$, where 
$B_t' = \post(l_a,B_t) \cap \gamma(A_t')$. The counterexample analysis procedure based on this annotation is given in Algorithm~\ref{algo:cex-analysis-safety}.
If each of the leaf nodes of the tree is annotated with a belief set $(l_a,B_t)$ for which $(l_a,B_t) \not\models p_k$, then the new annotation gives us a concrete counterexample tree $\counterex_\belief$, which by construction concertizes $\counterex_\abstr$. Conversely, if there exists a leaf node annotated with $(l_a,B_t)$ such that $(l_a,B_t) \models p_k$, then we can conclude that the abstract counterexample tree $\counterex_\abstr$ is not concretizable and use the path from the root of the tree to this leaf node to refine the partition $\part$.

\begin{algorithm}[b]
\vspace{-.4cm}
\small
\KwIn{surveillance game $(G,\LTLglobally p_k)$,\newline abstract counterexample tree $\counterex_\abstr$}
\KwOut{a path $\pi$ in $\counterex_\abstr$ or {\sc concretizable}}


\While{there is a node $v$ in $\counterex_\abstr$ whose children\newline are not annotated with states in $\states_\belief$}{
 let $(l_a,B_t)$ be the state with which $v$ is annotated\;
 \ForEach{child $v'$ of $v$ labelled with $(l_a',A_t')$}{
 annotate $v'$ with $(l_a',\post(l_a,B_t)\cap\gamma(A_t'))$\;
}
}%
\leIf{there is a path $\pi$ in $\counterex_\abstr$ from the root to a leaf annotated with a sate $s_\belief$ where $s_\belief\models p_k$\newline}
{\KwRet{$\pi$;}}
{\KwRet{{\sc concretizable}}}


\caption{Analysis of abstract counterexample trees for games with safety surveillance objectives.}
\label{algo:cex-analysis-safety}
\end{algorithm}

\begin{theorem}
If Algorithm~\ref{algo:cex-analysis-safety} returns a path $\pi_\abstr$ in $\counterex_\abstr$, then $\counterex_\abstr$ is not concretizable, and $\pi_\abstr$ is a non-concretizable path, otherwise  $\counterex_\abstr$ is concretizable.
\end{theorem}

\begin{figure}
\begin{center}
\begin{tikzpicture}[node distance=.7 cm,auto,>=latex',line join=bevel,transform shape,scale=.75]
\node at (0,0) (s0) {$v_0:(4,18)$};
\node  [below left of=s0,yshift=-.2cm,xshift=-2.2cm] (s1) {$v_1:(3,\{Q_2\})$};
\node  [below right of=s0,yshift=-.2cm,xshift=2.2cm] (s2) {$v_2:(9,\{Q_2\})$};
\node  [below of=s1,yshift=-.2cm] (s4) {$v_4:(4,A)$};
\node  [below of=s2,yshift=-.2cm] (s7) {$v_7:(8,A)$};
\node  [left of=s4,xshift=-1cm] (s3) {$v_3:(2,A)$};
\node  [right of=s4,xshift=1cm] (s5) {$v_5:(8,A)$};
\node  [left of=s7,xshift=-1cm] (s6) {$v_6:(4,A)$};
\node  [right of=s7,xshift=1cm] (s8) {$v_8:(14,A)$};
\draw [->] (s0) edge (s1.north);
\draw [->] (s0) edge (s2.north);
\draw [->] (s1) edge (s3.north);
\draw [->] (s1) edge (s4.north);
\draw [->] (s1) edge (s5.north);
\draw [->] (s2) edge (s6.north);
\draw [->] (s2) edge (s7.north);
\draw [->] (s2) edge (s8.north);

\end{tikzpicture}
\end{center}
\vspace{-.2cm}
\caption{Abstract counterexample in Example~\ref{ex:simple-safety-unconcretizable}. The leaf nodes are labelled with the abstract belief set $A = \{Q_1,Q_2\}$.}
\label{fig:simple-safety-counterex-1}
\vspace{-.5cm}
\end{figure}
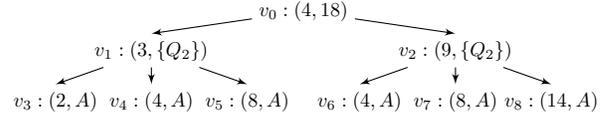

\begin{example}\label{ex:simple-safety-unconcretizable}
Let $G$ be the surveillance game structure from Example~\ref{ex:simple-surveillance-game}, and consider the surveillance game $(G,\LTLglobally p_5)$. 
Let $\part = \{Q_1,Q_2\}$ consist of the set $Q_1$, corresponding to the first two columns of the grid in Figure~\ref{simple-grid} and the set $Q_2$ containing the locations from the other three columns of the grid. Figure~\ref{fig:simple-safety-counterex-1} shows a counterexample tree $\counterex_\abstr$ in the abstract game $(\alpha_\part(G),\LTLglobally p_5)$. The analysis in Algorithm~\ref{algo:cex-analysis-safety} annotates node $v_1$ with the concrete belief set $\{17,23\}$, and the leaf node $v_3$ with the set $B = \{16,18,22,24\}$. Thus, this counterexample tree $\counterex_\abstr$ is determined to be unconcretizable and the partition $\part$ should be refined.\qed
\end{example}

When the analysis procedure determines that an abstract counterexample tree is unconcretizable, it returns a path in the tree that corresponds to a sequence of moves ensuring that the size of the belief-set does not actually exceed the threshold, given that the target behaves in a way consistent with the abstract counterexample.  Based on this path, we can then refine the abstraction in order to precisely capture this information and thus eliminate this abstract counterexample.

\subsubsection{Backward partition splitting}
Let $\pi_\abstr = v_0,\ldots, v_n$ be a path in $\counterex_\abstr$ where $v_0$ is the root node and $v_n$ is a leaf. For each node $v_i$, let $(l_a^i,A_t^i) $ be the abstract state labelling $v_i$ in $\counterex_\abstr$, and let $(l_a^i,B_t^i)$ be the  belief set with which the node was annotated by the counterexample analysis procedure. We consider the case when $(l_a^n,B_t^n) \models p_k$, that is, $|\{l_t \in B_t^n \mid \vis(l_a,l_t) = \false\}| \leq k$.
Note that since $\counterex_\abstr$ is a counterexample we have $(l_a^n,A_t^n) \not \models p_k$, and since $k>0$, this means $A_t \in \mathcal{P}(\part)$.

We now describe a procedure to compute a partition $\part'$ that refines the current partition $\part$ based on the path $\pi_\abstr$. Intuitively, we split the sets that appear in $A_t^n$ in order to ensure that in the refined abstract game the corresponding abstract state satisfies the surveillance predicate $p_k$. We may have to also split sets appearing in abstract states on the path to $v_n$, as we have to ensure that earlier imprecisions on this path do not propagate, thus including more than the desired newly split sets, and leading to the same violation of $p_k$.

Formally, if $A_t^n = (l_a^n,\{B_{n,1},\ldots,B_{n,m_n}\})$, then we split some of the sets $B_{n,1},\ldots,B_{n,m_n}$ to obtain from $A_t^n$ a set $A'_n = \{B_{n,1}',\ldots,B_{n,m_n'}'\}$ such that

$\qquad|\{l_t \in \gamma(C^n) \mid \vis(l_a^n,l_t) = \false\}| \leq k \text{, where}$

$\qquad\qquad C^n = \{B_{n,i}' \in A'_n \mid B_{n,i}' \cap B_t^n \neq \emptyset\}.$

This property intuitively means that if we consider the sets in $A'_n$ that have non-empty intersection with $B_t^n$, an abstract state composed of those sets will satisfy $p_k$. Since $(l_a^n,B_t^n)$ satisfies $p_k$, we can find a partition $\part^n \preceq \part$ that guarantees this property, as shown in Algorithm~\ref{algo:refinement-safety}.
 
What remains, in order to eliminate this counterexample, is to ensure that only these sets are reachable via the considered path, by propagating this split backwards, obtaining a sequence of partitions $\part \succeq \part^n \succeq \ldots \succeq \part^0$ refining $\part$. 

Given $\part^{j+1}$, we compute $\part^j$ as follows. For each $j$, we define a set $C^j \subseteq \mathcal{P}(L_t)$ (for $j=n$, the set $C^n$ was defined above). Suppose we have defined $C^{j+1}$ for some $j \geq 0$, and $A_t^j = (l_a^j,\{B_{j,1},\ldots,B_{j,m_j}\})$. We split some of the sets $B_{j,1},\ldots,B_{j,m_j}$ to obtain from $A_t^j$ a set $A'_j = \{B_{j,1}',\ldots,B_{j,m_j'}'\}$ where there exists $C^j \subseteq A'_j$ with
\[\gamma(C^j) = \{l_t \in \gamma(A_t^j) \mid \post(l_a^j,\{l_t\}) \cap \gamma(A_t^{j+1}) \subseteq \gamma(C^{j+1})\}.\]
This means that using the new partition we can express precisely the set of states that do not lead to sets in $A_{j+1}'$ that we are trying to avoid. 
The fact that an appropriate partition $\part$ can be computed, follows from the choice of the leaf node $v_n$. 
The procedure for computing the partition $\part' = \part^0$ that refines $\part$ based on  $\pi_\abstr$ is formalized in Algorithm~\ref{algo:refinement-safety}.
\begin{example}
We continue with the unconcretizable abstract counterexample tree from Example~\ref{ex:simple-safety-unconcretizable}. We illustrate the refinement procedure for the path $v_0,v_1,v_3$. For node $v_3$, we split $Q_1$ and $Q_2$ using the set $B = \{16,18,22,24\}$, obtaining the sets $Q_1' = Q_1 \cap \{16,18,22,24\} = \{16\}$, $Q_2' = Q_1\setminus\{16\}$, $Q_3' = Q_2 \cap \{16,18,22\} = \{18,22,24\}$ and $Q_4' = Q_2 \setminus \{18,22,24\}$. We thus obtain a new partition $\part_{v_3} \preceq \part$. In order to propagate the refinement backwards (to ensure eliminating $\counterex_\abstr$) we compute the set of locations in $Q_2$ from which the target can move to a location in $Q_2'$ or $Q_4'$ that is not visible from location $3$. In this case, these are just the locations $18$, $22$ and $24$, which have already been separated from $Q_2$, so here backward propagation does not require further splitting.\qed
\end{example}

\begin{algorithm}[b]
\vspace{-.4cm}
\small
\KwIn{surveillance game $(G,\LTLglobally p_k)$, abstraction partition $\part$,\newline unconcretizable path $\pi = v_0,\ldots,v_n$ in $\counterex_\abstr$}
\KwOut{an abstraction partition $\part'$ such that $\part' \preceq \part$}


let $(l_a^j,A_t^j)$ be the label of $v_j$, and $(l_a^j,B_t^j)$ its annotation;
\begin{flalign*}
A  :=&  \{Q \cap B_t^n\mid Q \in A_t^n , Q \cap B_t^n \neq \emptyset \}\cup &\\
& \{Q \setminus B_t^n\mid Q \in A_t^n , Q \setminus B_t^n \neq \emptyset \};
\end{flalign*}

$\part' := (\part \setminus A_t^n )  \cup A$;
$C := \{ Q\in A \mid Q \cap B_t^n \neq \emptyset\}$

\For{$j = n-1,\ldots,0$}{
\lIf{$A_t^j \in L_t$}{{\bf break}}
$B :=  \{l_t \in \gamma(A_t^j) \mid \post(l_a,\{l_t\}) \subseteq \gamma(C)\}$\;
\noindent
\begin{flalign*}
A  := &\{Q \cap B\mid Q \in A_t^j , Q \cap B \neq \emptyset \}\cup &\\
& \{Q \setminus B\mid Q \in A_t^j , Q \setminus B \neq \emptyset \};
\end{flalign*}

$\part' := (\part' \setminus A_t^j )  \cup A$;
$C := \{ Q\in A \mid Q \cap B \neq \emptyset\}$
} 
\KwRet{$\part'$}


\caption{Abstraction partition refinement given an unconcretizable path in an  abstract counterexample tree.}
\label{algo:refinement-safety}

\end{algorithm}

Let $\part$ and $\part'$ be two counterexample partitions such that $\part' \preceq \part$. Let $\counterex_\abstr$ be an abstract counterexample  tree in $(\alpha_\part(G),\LTLglobally p_k)$. We define $\gamma_{\part'}(\counterex_\abstr)$ to be the set of abstract counterexample trees in $(\alpha_{\part'}(G),\LTLglobally p_k)$ such that $\counterex'_\abstr \in \gamma_{\part'}(\counterex_\abstr)$ iff $\counterex'_\abstr$ differs from $\counterex_\abstr$ only in the node labels and for every node in $\counterex_\abstr$ labelled with $(l_a,A_t)$, the corresponding node in $\counterex'_\abstr$ is labelled with an abstract state $(l_a,A_t')$ such that $\gamma(A_t') \subseteq \gamma(A_t)$.
 
The theorem below states the progress property (eliminating the considered counterexample) of Algorithm~\ref{algo:refinement-safety}.

\begin{theorem}If $\part'$ is the partition returned by Algorithm~\ref{algo:refinement-safety} for an unconcretizable abstract counterexample $\counterex_\abstr$ in $(\alpha_\part(G),\LTLglobally p_k)$, then $\gamma_{\part'}(\counterex_\abstr) = \emptyset$, and also $\gamma_{\part''}(\counterex_\abstr) = \emptyset$ for every partition $\part''$ where $\part'' \preceq \part'$.
\end{theorem}

\begin{example}\label{ex:simple-safety-realizability}
In the surveillance game $(G,\LTLglobally p_5)$, where $G$ is the surveillance game structure from Example~\ref{ex:simple-surveillance-game}, the agent has a winning strategy. After $6$ iterations of the refinement loop we arrive at an abstract game $(\alpha_{\part^*}(G),\LTLglobally p_5)$, where the partition $\part^*$ consists of $11$ automatically computed sets (as opposed to the $22$ locations reachable by the target in $G$), which in terms of the belief-set construction means $2^{11}$ versus $2^{22}$ possible belief sets in the respective games.

In the game $(G,\LTLglobally p_2)$, on the other hand, the agent does not have a winning strategy, and our algorithm establishes this after one refinement, after which, using a partition of size $4$,  it finds a concretizable abstract counterexample.
\qed
\end{example}


\section{BELIEF REFINEMENT FOR LIVENESS}
\subsection{Counterexample Graph}
The counterexamples for general surveillance properties are directed graphs, which may contain cycles. In particular, for a liveness surveillance property of the form $\LTLglobally\LTLfinally p_k$ each infinite path in the graph has a position such that, from this position on, each state on the path violates $p_k$. An \emph{abstract counterexample graph} in the game $(G_\abstr,\LTLglobally\LTLfinally p_k)$ is a finite graph $\counterex_\abstr$ defined analogously to the abstract counterexample tree. The difference being that there are no leaves, and that for each cycle $\rho = v_1,v_2,\ldots,v_n$ with $v_1 = v_n$ in $\counterex_\abstr$ that is reachable from $v_0$, every node $v_i$ in $\rho$ is labelled with state $s_\abstr^i$ where $s_\abstr^i \not\models p_k$.

\begin{figure}
\begin{minipage}{0.2\textwidth}
\begin{center}
\begin{tikzpicture}[scale=0.9]
\draw[step=0.5cm,color=gray] (-1.5,-1.5) grid (1,1);
\filldraw[fill=blue,draw=black] (+0.75,+0.75) circle (0.2cm);
\filldraw[fill=red,draw=black] (0,0) rectangle (-0.5,-0.5);
\filldraw[fill=red,draw=black] (-0.5,0) rectangle (-1,-0.5);
\filldraw[fill=red,draw=black] (0,0) rectangle (0.5,-0.5);
\filldraw[fill=blue!40!white,draw=black] (+0.75,+0.75) circle (0.2cm);
\filldraw[fill=orange!40!white,draw=black] (0.25,-0.75) circle (0.2cm);
\draw[blue,thick] (-1.35,-0.75) rectangle (0.75,0.25);
\draw[blue,thick,->] (0.73,0.75) -> (0.75,0.25);
\node at (-1.30,+0.75) {\tiny{0}};
\node at (-0.80,+0.75) {\tiny{1}};
\node at (-0.30,+0.75) {\tiny{2}};
\node at (0.20,+0.75) {\tiny{3}};
\node at (0.73,+0.75) {\tiny{4}};
\node at (-1.35,+0.25) {\tiny{5}};
\node at (-0.85,+0.25) {\tiny{6}};
\node at (-0.35,+0.25) {\tiny{7}};
\node at (0.25,+0.25) {\tiny{8}};
\node at (0.75,+0.25) {\tiny{9}};
\node at (-1.35,-0.25) {\tiny{10}};
\node at (-0.85,-0.25) {\tiny{11}};
\node at (-0.35,-0.25) {\tiny{12}};
\node at (0.25,-0.25) {\tiny{13}};
\node at (0.75,-0.25) {\tiny{14}};
\node at (-1.35,-0.75) {\tiny{15}};
\node at (-0.85,-0.75) {\tiny{16}};
\node at (-0.35,-0.75) {\tiny{17}};
\node at (0.25,-0.75) {\tiny{18}};
\node at (0.75,-0.75) {\tiny{19}};
\node at (-1.35,-1.25) {\tiny{20}};
\node at (-0.85,-1.25) {\tiny{21}};
\node at (-0.35,-1.25) {\tiny{22}};
\node at (0.25,-1.25) {\tiny{23}};
\node at (0.75,-1.25) {\tiny{24}};
\end{tikzpicture}
\end{center}
\end{minipage}
\begin{minipage}{0.28\textwidth}
\vspace{0.1cm}
\caption{\small Agent locations on an (infinite) path in the abstract counterexample graph from Example~\ref{ex:simple-liveness-counterex}. In the graph, the first node is labelled with $(4,18)$, the second with $(9,\{Q_2\})$, and all other nodes with some $(l_a,\{Q_1,Q_2\})$.}
\label{fig:simple-liveness-counterex}
\end{minipage}
\vspace{-.65cm}
\end{figure}
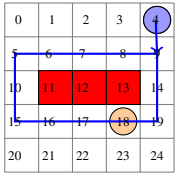

\begin{example}\label{ex:simple-liveness-counterex}
We saw in Example~\ref{ex:simple-safety-realizability} that in the safety surveillance game $(G,\LTLglobally p_2)$ the agent does not have a winning strategy. 
We now consider a relaxed requirement, namely, that the uncertainty drops to at most $2$ infinitely often. We consider the liveness surveillance game 
$(G,\LTLglobally \LTLfinally p_2)$.

Let $\part = \{Q_1,Q_2\}$ be the partition from Example~\ref{ex:simple-safety-unconcretizable}. 
Figure~\ref{fig:simple-liveness-counterex} shows an infinite path (in lasso form) in the abstract game $(\alpha_\part(G),\LTLglobally \LTLfinally p_2)$.  The figure depicts only the corresponding trajectory (sequence of positions) of the agent. The initial abstract state is $(4,18)$, the second node on the path is labeled with the abstract state $(9,\{Q_2\})$, and all other nodes on the path are labeled with abstract states of the form $(l_a,\{Q_1,Q_2\})$. As each abstract state in the cycle violates $p_2$, the path violates $\LTLglobally \LTLfinally p_2$. The same holds for all infinite paths in the existing abstract counterexample graph.
\qed
\end{example}

A \emph{concrete counterexample graph} $\counterex_\belief$ for the belief game $(G_\belief,\LTLglobally\LTLfinally p_k)$ is defined analogously. 

An abstract counterexample graph $\counterex_\abstr$ for the game $(G_\abstr,\LTLglobally\LTLfinally p_k)$ is \emph{concretizable} if there exists a counterexample
$\counterex_\belief$ in $(G_\belief,\LTLglobally \LTLfinally p_k)$, such that for each infinite path $\pi_\abstr = v_\abstr^0,v_\abstr^1,\ldots$ starting from the initial node of $\counterex_\abstr$ there exists an infinite path $\pi_\belief = v_\belief^0,v_\belief^1,\ldots$ in $\counterex_\belief$ staring from its initial node such that if $v_\abstr^i$ is labelled with $(l_a,A_t)$ in $\counterex_\abstr$, then the corresponding node $v_\belief^i$ in $\counterex_\belief$ is labelled with $(l_a,B_t)$ for some $B_t \in \mathcal{P}(L_t)$ for which $B_t \subseteq \gamma(A_t)$.

\subsection{Counterexample-Guided Refinement}
\subsubsection{Forward belief-set propagation}

To check if an abstract counterexample graph $\counterex_\abstr$ is concretizable, we construct a finite graph $\mathcal{D}$ whose nodes are labelled with elements of $\states_\belief$ and with nodes of $\counterex_\abstr$.
By construction we will ensure that if a node $d$ in $\mathcal D$ is labelled with $\langle(l_a,B_t),v \rangle$, where $(l_a,B_t)$ is a belief state, and $v$ is a node in $\counterex_\abstr$, then $v$ is labelled with $(l_a,A_t)$ in $\counterex_\abstr$, and $B_t \subseteq \gamma(A_t)$. 

Initially $\mathcal D$ contains a single node $d_0$ labelled with $\langle s_\belief^\init,v_0\rangle$, where $v_0$ is initial node of $\counterex_\abstr$. Consider a node $d$ in $\mathcal D$ labelled with $\langle(l_a,B_t),v \rangle$. For every child $v'$ of $v$ in $\counterex_\abstr$, labelled with an abstract state $(l_a',A_t')$ we proceed as follows. We let ${B_t}' = \post(l_a,B_t) \cap \gamma(A_t')$. If there exists a node $d'$ in $\mathcal D$ labelled with $\langle (l_a',B_t'),v\rangle$, then we add an edge from $d$ to $d'$ in $\mathcal{D}$. Otherwise, we create such a node and add the edge. We continue until no more nodes and edges can be added to $\mathcal D$. The procedure is guaranteed to terminate, since both  the graph $\counterex_\belief$, and $\states_\belief$ are finite, and we add a node labelled $\langle s_\belief, v\rangle$ to $\mathcal D$ at most once.

If the graph $\mathcal D$ contains a reachable cycle (it suffices to consider simple cycles, i.e., without repeating intermediate nodes) $\rho = d_0,\ldots,d_n$ with $d_0 = d_n$ such that some $d_i$ is labelled with $(l_a,B_t)$ where $(l_a,B_t) \models p_k$, then we conclude that the abstract counterexample $\counterex_\abstr$ is not concretizable. If no such cycle exists, then $\mathcal D$ is a concrete counterexample graph for the belief game $(G_\belief,\LTLglobally\LTLfinally p_k)$. 

\begin{algorithm}[b] \small
\vspace{-.5cm}
\KwIn{surveillance game $(G,\LTLglobally\LTLfinally p_k)$, abstract counterexample graph $\counterex_\abstr$ with initial node $v_0$}
\KwOut{a path $\pi$ in a graph $\mathcal D$ or {\sc concretizable}}

\smallskip

graph $\mathcal D = (D,E)$ with nodes $D := \{d_0\}$ and edges $E := \emptyset$\;

annotate $d_0$ with $\langle s_\belief^\init, v_0\rangle$\; 

\SetKwRepeat{Do}{do}{while}

\Do{$\mathcal D \neq \mathcal D'$}{
$\mathcal D' := \mathcal D$\;
 \ForEach{node $d$ in $\mathcal D$ labelled with $\langle(l_a,B_t),v\rangle$}{
  \ForEach{child $v'$ of $v$ in $\counterex_\abstr$ labelled $(l_a',A_t')$}{
  $B_t' := \post(l_a,B_t)\cap\gamma(A_t')$\;
  \eIf{there is a node $d' \in D$ labelled with $\langle (l_a',B_t'),v'\rangle$}{add an edge $(d,d')$ to $E$}
  {add a node $d'$ labelled $\langle (l_a',B_t'),v'\rangle$ to $D$\;
  add an edge $(d,d')$ to $E$}
}
}
}
\leIf{there is a lasso path $\pi$ in $\mathcal D$ starting from $d_0$ such that some node in the cycle is annotated with $\langle s_\belief,v\rangle$ and $s_\belief\models p_k$\newline}
{\KwRet{$\pi$;}}
{\KwRet{{\sc concretizable}}}

\smallskip

\caption{Analysis of abstract counterexample graphs for games with liveness surveillance objectives.}
\label{algo:cex-analysis-liveness}
\end{algorithm}

\begin{example}\label{ex:simple-liveness-unconcretizable}
The abstract counterexample graph in the game $(\alpha_\part(G),\LTLglobally \LTLfinally p_2)$ discussed in Example~\ref{ex:simple-liveness-counterex} is not conretizable, since for the path in the abstract graph depicted in Figure~\ref{fig:simple-liveness-counterex} there exists a corresponding path in the graph $\mathcal D$ with a node in the cycle labelled with a set in $G_\belief$ that satisfies $p_2$. More precisely, the cycle in the graph $\mathcal D$ contains a node labelled with $(19,\{10\})$. Intuitively, as the agent moves from the upper to the lower part of the grid along this path, upon not observing the target, it can infer from the sequence of observations that the only possible location of the target is $10$. Thus, this paths is winning for the agent.
\qed
\end{example}

\begin{theorem}
If Algorithm~\ref{algo:cex-analysis-liveness} returns a path $\pi$ in the graph $\mathcal D$ constructed for $\counterex_\abstr$, then $\counterex_\abstr$ is not concretizable, and the infinite run in $G_\belief$ corresponding to $\pi$ satisfies $\LTLglobally\LTLfinally p_k$, otherwise  $\counterex_\abstr$ is concretizable.
\end{theorem}

\subsubsection{Backward partition splitting}

Consider a path in the graph $\mathcal{D}$ of the form $\pi = d_0,\ldots, d_n,d_0',\ldots,d_m'$ where $d_n = d_m'$, and where for some $0 \leq i \leq m$ for the label $(l_a^i,B_t^i)$ it holds that $(l_a^i,B_t^i) \models p_k$. Let 
$\pi_\abstr = v_0,\ldots, v_n,v_0',\ldots,v_m'$ be the sequence of nodes in $\counterex_\abstr$ corresponding to the labels in $\pi$. By construction of $\mathcal D$, $\pi_\abstr$ is a path in $\counterex_\abstr$ and $v_n = v_m'$. We apply the refinement procedure from the previous section to the whole path $\pi_\abstr$, as well as to the path-prefix $v_0,\ldots, v_n$.

Let $\part$ and $\part'$ be two counterexample partitions such that $\part' \preceq \part$. Let $\counterex_\abstr$ be an abstract counterexample graph in $(\alpha_\part(G),\LTLglobally\LTLfinally p_k)$. We define $\gamma_{\part'}(\counterex_\abstr)$ to be the set of abstract counterexample graphs in $(\alpha_{\part'}(G),\LTLglobally\LTLfinally p_k)$ such that for every infinite path $\pi$ in $\counterex_\abstr'$ there exists an infinite path $\pi$ in $\counterex_\abstr$ such that for every node in $\pi'$ labelled with $(l_a,A_t')$ the corresponding node in $\counterex_\abstr$ is labelled with an abstract state $(l_a,A_t)$ such that $\gamma(A_t') \subseteq \gamma(A_t)$.

\begin{theorem}If $\part'$ is the partition 
obtained by refining $\part$ with respect to an uncocretizable abstract counterexample $\counterex_\abstr$ in $(\alpha_\part(G),\LTLglobally\LTLfinally p_k)$, then $\gamma_{\part'}(\counterex_\abstr) = \emptyset$, and also $\gamma_{\part''}(\counterex_\abstr) = \emptyset$ for every partition $\part''$ with $\part'' \preceq \part'$.
\end{theorem}

\begin{example}\label{ex:simple-liveness-refinement}
We refine the abstraction partition $\part$ from Example~\ref{fig:simple-liveness-counterex} using the path identified there, in order to eliminate the abstract counterexample. For this, following the refinement algorithm, we first split the set $Q_1$ into sets $Q_1' = \{10\}$ and $Q_2' = Q_1 \setminus \{10\}$, and let $Q_3' = Q_2$. However, since from some locations in $Q_2'$ and in $Q_3'$ the target can reach locations in $Q_2'$ and $Q_3'$ that are not visible from the agent's position $19$, in order to eliminate the counterexample, we need to propagate the refinement backwards along the path and split $Q_2'$ and $Q_3'$ further. With that, we obtain an abstraction partition with $10$ sets, which is guaranteed to eliminate this abstract counterexample. In fact, in this example this abstraction turns out to be sufficiently precise to obtain a winning strategy for the agent.
\qed
\end{example}

\subsection{General surveillance and task specifications}
We have described refinement procedures for safety and liveness surveillance objectives. If we are given a conjunction of such objectives, we first apply the refinement procedure for safety, and if no path for which we can refine is found, we then apply the refinenment procedure for liveness. 

In the general case, we check if the counterexample contains a state for which the concrete belief is a strict subset of the abstract one. If this is not the case, then the counterexample is concretizable, otherwise we refine the abstraction to make this belief precise. In the special case when we have a conjunction of a surveillance and task specifications, we first refine with respect to the surveillance objective as described above, and if this is not possible, with respect to such a node. Since the set of states in the game is finite, the iterative refinement will terminate, either with a concretizable counterexample, or with a surveillance strategy.



\section{EXPERIMENTAL EVALUATION}\label{sec:experiments}
We report on the application of our method for surveillance synthesis in two case studies. We have implemented the simulation in \texttt{Python}, using the \texttt{slugs} reactive synthesis tool~\cite{EhlersR16}. The experiments were performed on an Intel i5-5300U 2.30 GHz CPU with 8 GB of RAM. 

\subsection{Liveness Surveillance Specification + Task Specification}
\begin{figure}
\subfloat[Gridworld with a user provided abstraction partition with 7 sets, marked by black lines.. \label{fig:case1}]{
\includegraphics[scale=0.3]{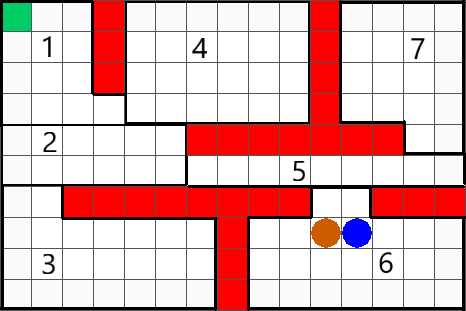}
}
\hfill
\subfloat[Gridworld showing visibility of the agent. All locations shown in black are invisible to the agent. \label{fig:case1vis}]{
\includegraphics[scale=0.3]{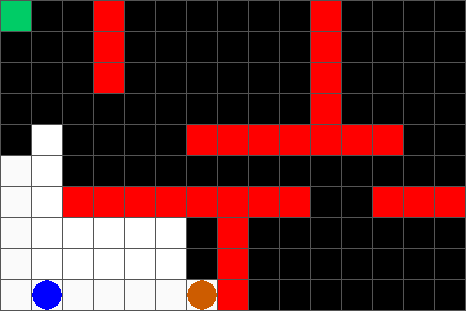}\hspace{.5cm}
}

\caption{10x15 gridworld with a surveillance liveness specification. The agent is blue, and the target to be surveilled is orange. Red states are obstacles.}
\label{fig:casestudies}

\end{figure}

Figure~\ref{fig:case1} shows a gridworld divided into  regions. The surveillance objective requires the agent to infinitely often know precisely the location of the target (either see it, or have a belief consisting of one cell). Additionally, it has to perform the task of patrolling (visiting infinitely often) the green 'goal' cell. Formally, the specification is $\LTLglobally\LTLfinally p_1 \wedge \LTLglobally\LTLfinally \mathit{goal}$. The agent can move up to 3 grid cells away at each step. The sensor mode, that is, the visibility function, used here is 'line-of-sight' with a range of 5 cells. The agent cannot see through obstacles (shown in red) and cannot see farther than 5 cells away.

Using the abstraction partition of size 7 shown in Figure \ref{fig:case1}, the overall number of states in the two-player game is $15\times10 + 2^7 = 278$ states. In contrast, solving the full abstract game will have in the order of $2^{150}$ states, which is a state-space size that state-of-the-art synthesis tools cannot handle. 

\begin{figure}
\begin{minipage}{5.0cm}
	\centering
		\subfloat[$t_1$ \label{fig:case1t2}]{
		\includegraphics[scale=0.17]{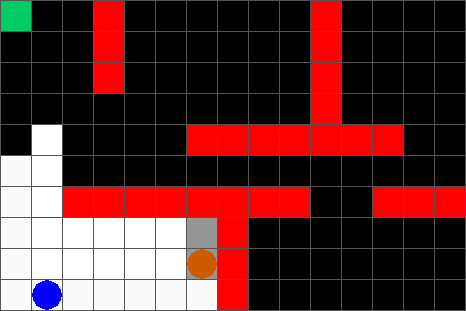}\hspace{.5cm}
	}
	\subfloat[$t_3$ \label{fig:case1t3}]{
		\includegraphics[scale=0.17]{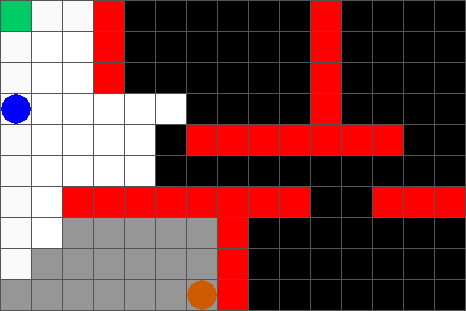}\hspace{.5cm}
	}
	\subfloat[$t_4$ \label{fig:case1t4}]{
	\includegraphics[scale=0.17]{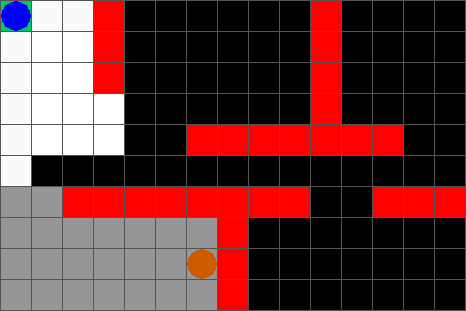}\hspace{.5cm}
}
\end{minipage}
\begin{minipage}{5.0cm}
	\centering
	\subfloat[$t_5$  \label{fig:case1t5}]{
		\includegraphics[scale=0.17]{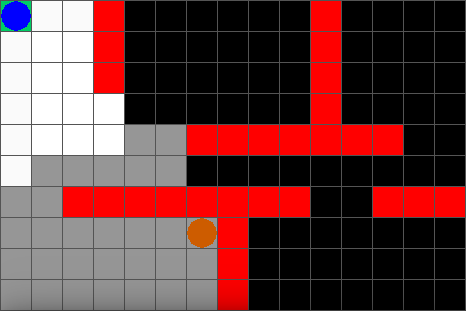}\hspace{.5cm}
	}
	\subfloat[$t_6$ \label{fig:case1t6}]{
		\includegraphics[scale=0.17]{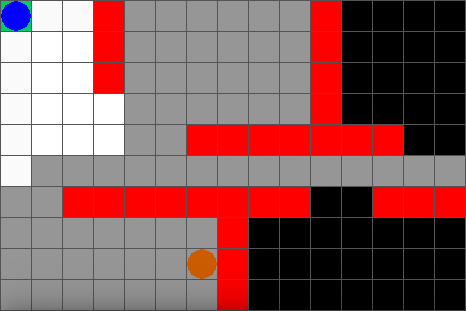}\hspace{.5cm}
	}
	\subfloat[$t_7$ \label{fig:case1t7}]{
		\includegraphics[scale=0.17]{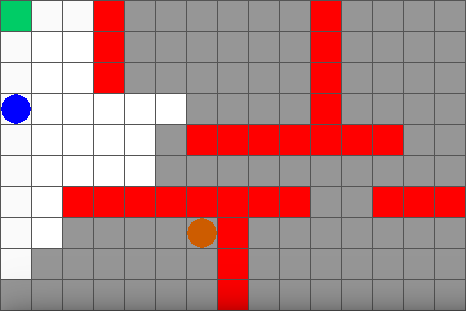}\hspace{.5cm}
	}
	
\end{minipage}

	\caption{Evolution of the agent's belief about the target's location as it moves to the goal and loses sight of the target. Grey cells represent the locations the agent believes the target could be in. We show the belief at different timesteps $t_1,\ldots,t_7$ (note that $t_2$ is excluded for space concerns)
		}
	\label{fig:case1exp}
	
\end{figure}

\begin{figure}
	\begin{minipage}{5.0cm}
		\centering
		\subfloat[$t_5$ \label{fig:search1}]{
			\includegraphics[scale=0.17]{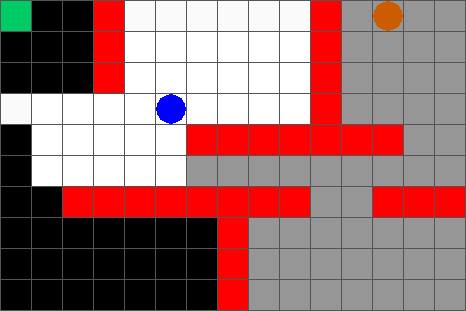}\hspace{.5cm}
		}
		\subfloat[$t_7$ \label{fig:search2}]{
			\includegraphics[scale=0.17]{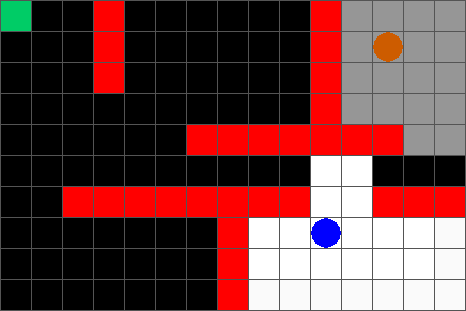}\hspace{.5cm}
		}
		\subfloat[$t_9$ \label{fig:search3}]{
			\includegraphics[scale=0.17]{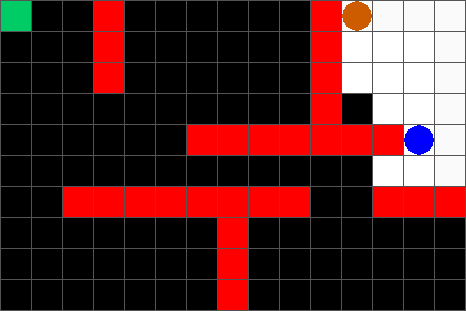}\hspace{.5cm}
		}
	\end{minipage}
	\caption{The agent has to search for the target in order to lower its belief below the surveillance liveness specification.
	}
	\label{fig:search}
	
\end{figure}

Figure \ref{fig:case1exp} shows how the belief of the agent (shown in grey) can grow quickly when it cannot see the target. This growth occurs due to the coarseness of the abstraction, which overapproximates the target's true position. In 7 steps, the agent believes the target can be anywhere in the grid that is not in its vision. It has to then find the target in order to satisfy the surveillance requirement. Figure \ref{fig:search} illustrates the searching behaviour of the agent when it is trying to lower the belief below the threshold in order to satisfy the liveness specification. The behaviour of the agent shown here will contrast with the behaviour under safety surveillance which will we look at next.

\begin{figure}
	\begin{minipage}{5.0cm}
		\centering
		\subfloat[$t_1$ \label{fig:casefine1t2}]{
			\includegraphics[scale=0.17]{figs/Liveness_t1.png}\hspace{.5cm}
		}
		\subfloat[$t_3$ \label{fig:case1finet3}]{
			\includegraphics[scale=0.17]{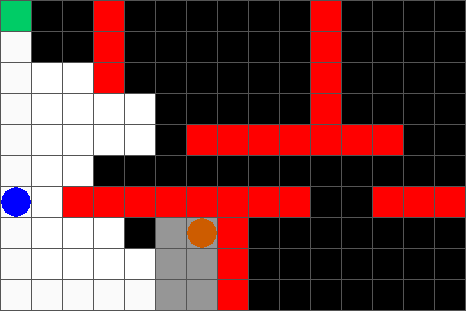}\hspace{.5cm}
		}
		\subfloat[$t_4$ \label{fig:case1finet4}]{
			\includegraphics[scale=0.17]{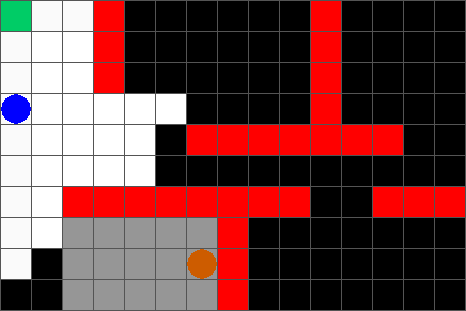}\hspace{.5cm}
		}
	\end{minipage}
	\begin{minipage}{5.0cm}
		\centering
		\subfloat[$t_5$  \label{fig:case1finet5}]{
			\includegraphics[scale=0.17]{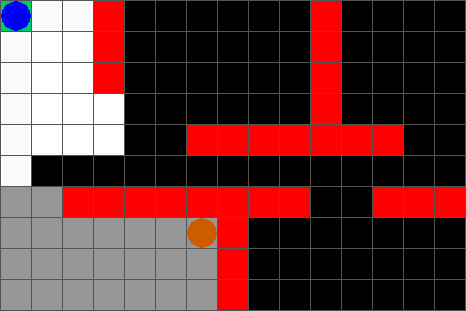}\hspace{.5cm}
		}
		\subfloat[$t_6$ \label{fig:case1finet6}]{
			\includegraphics[scale=0.17]{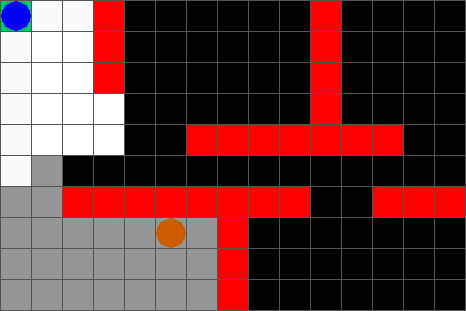}\hspace{.5cm}
		}
		\subfloat[$t_7$ \label{fig:case1finet7}]{
			\includegraphics[scale=0.17]{figs/Liveness_t5.png}\hspace{.5cm}
		}
		
	\end{minipage}

	\caption{Evolution of the agent's belief about the target's location in a game  with an abstraction partition of size 12.
	}
	\label{fig:case1fineexp}
	
\end{figure}

In this example, an abstraction partition of size 7 was enough to guarantee the satisfaction surveillance specification.  For the purpose of comparison, we also solve the game with  an abstraction partition of size 12  to illustrate the change in belief growth. Figure \ref{fig:case1fineexp} shows the belief states growing much more slowly as the abstract belief states are smaller, and thus they more closely  approximate the true belief of the agent.

The additional abstraction partitions result in a much larger game as the state space grows exponentially in the size of the abstraction partition. Table \ref{tab:exp1} compares the sizes of the corresponding abstract games, and the time it takes to synthesize a surveillance controller in each case.

\begin{table}[h!]
	\centering
	\begin{tabular}{c|c|c}
	Size of abstraction partition & Size of abstract game & Synthesis time \\ \hline \hline
		7 & 278 & 237s \\ 
		12 & 4346 & 810s \\ 
	\end{tabular}\caption{Comparison of synthesis times for the two cases} \label{tab:exp1}
\end{table}

A video simulation can be found at \url{http://goo.gl/YkFuxr}.

\subsection{Safety surveillance specification + task specification}
\begin{figure}
\centering
\includegraphics[scale=0.13]{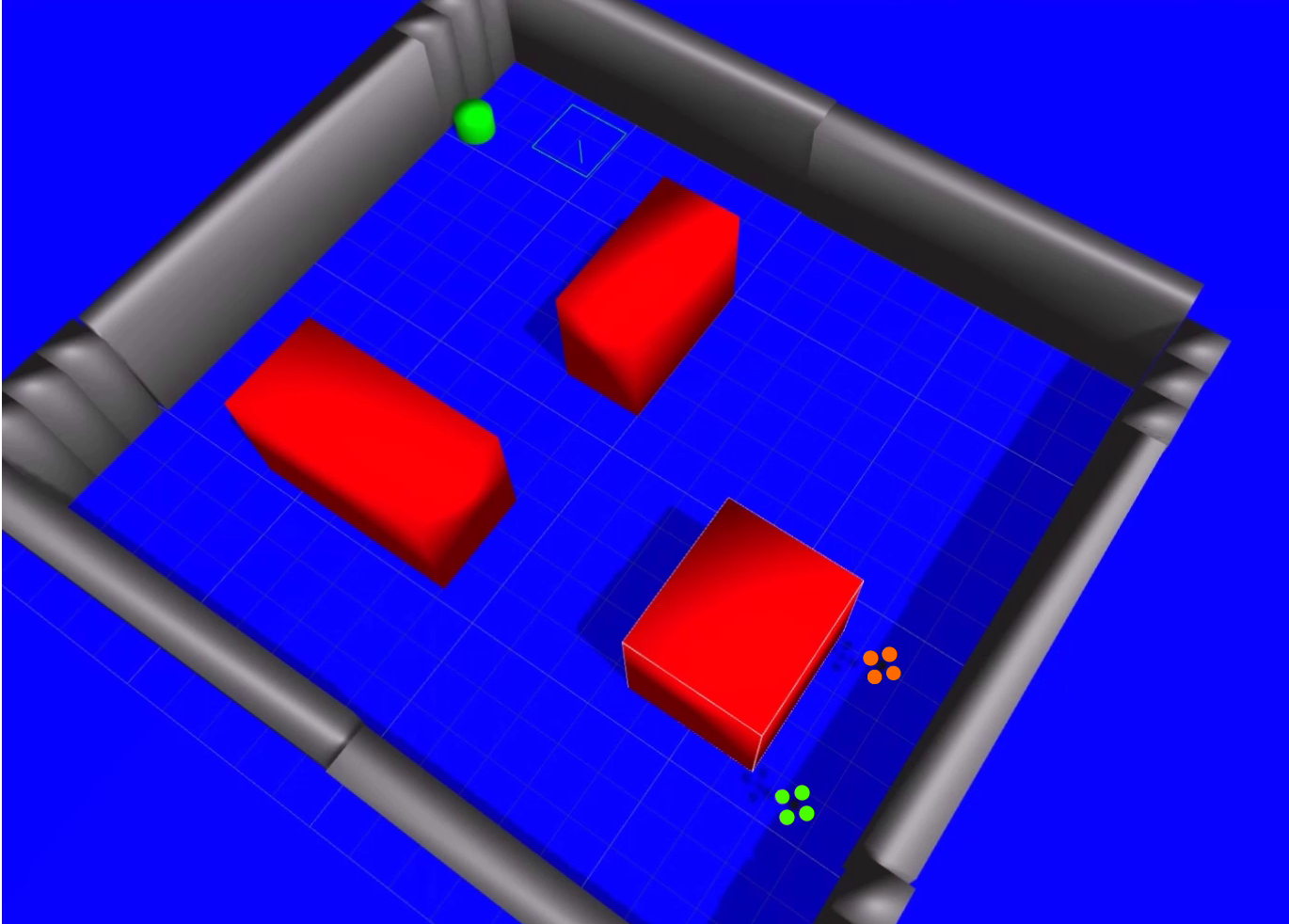}\caption{A Gazebo environment where the red blocks are obstacles that the drones cannot see past. The green drone is the agent and the orange drone is the target.}\label{fig:case2}
\vspace{-.5cm}
\end{figure}
Figure~\ref{fig:case2} depicts an environment created in \emph{Gazebo} where the red blocks model buildings. The drone is given full line of sight vision - it can detect the target if there is no obstacle in the way. 
In this setting, we enforce the safety surveillance objective $\square p_{30}$ (the belief size should never exceed 30) in addition to infinitely often reaching the green cell. The formal specification is $\LTLglobally\LTLfinally p_{30} \wedge \LTLglobally\LTLfinally \mathit{goal}$. Additionally, the target itself is trying to reach the goal cell infinitely often as well, which is known to the agent.

We used an abstraction generated by a partition of size 6, which was sufficiently precise to compute a surveillance strategy in 210 s. Again, note that the precise belief-set game would have in the order of $2^{200}$ states.
 
We simulated the environment and the synthesized surveillance strategy for the agent in Gazebo and ROS. A video of the simulation can be found at \url{http://goo.gl/LyC1gQ}. This simulation presents a qualitative difference in behaviour compared to the previous example. There, in the case of liveness surveillance, the agent had more leeway to completely lose the target in order to reach its goal location, even though the requirement of reducing the size of the belief to $1$ is quite strict. Here, on the other hand, the safety surveillance objective, even with a large threshold of $30$, forces the agent to follow the target more closely, in order to prevent its belief from getting too large. The synthesis algorithm thus provides the ability to obtain qualitatively different behaviour as necessary for specific applications by combining different objectives. 

\subsection{Discussion}
The difference in the behaviour in the case studies highlights the different use cases of the surveillance objectives. Depending on the domain, the user can specify a combination of safety and liveness specification to tune the behaviour of the agent. In a critical surveillance situation (typical in defense or security situations), the safety specification will guarantee to the user that the belief will never grow too large. However, in less critical situations (such as luggage carrying robots in airports), the robot has more flexibility in allowing the belief to grow as long as it can guarantee its reduction in the future. 


\section{CONCLUSIONS}
We have presented a novel approach to solving a surveillance problem with information guarantees. We provided a framework that enables the  formalization of the surveillance synthesis problem as a two-player, partial-information game. We then presented a method to reason over the belief that the agent has over the target's location and specify formal surveillance requirements. The user can tailor the behaviour to their specific application by using a combination of safety and liveness surveillance objectives.

The benefit of the proposed framework is that it allows it leverages techniques successfully used in verification and reactive synthesis to develop efficient methods for solving the surveillance problem. There are several promising  avenues of future work using and extending this framework. Some of which currently being explored are the following;
\begin{itemize}
\item Synthesizing distributed strategies for multi-agent surveillance in a decentralized manner. These compositional synthesis methods avoid the blow up of the state space that occurs in centralized synthesis procedures as the number of surveillance agents grow.
\item Incorporating static sensors or alarm triggers for the mobile agent(s) to coordinate with.
\item Allowing for sensor models to include uncertainty and detection errors while still providing surveillance guarantees.

\end{itemize}





\bibliographystyle{IEEEtran}
\bibliography{main}

\end{document}